\pdfoutput=1

\documentclass[11pt]{article}

\usepackage{ACL}

\usepackage{times}
\usepackage{latexsym}

\usepackage[T1]{fontenc}

\usepackage[utf8]{inputenc}

\usepackage{microtype}

\usepackage{inconsolata}

\usepackage{amsmath}
\usepackage{amsfonts}
\usepackage{graphicx}
\usepackage{subcaption}
\usepackage{tikz}
\usepackage{multirow}
\usepackage{float}
\usepackage{booktabs}
\usepackage{tcolorbox}
\usepackage{float}
\usepackage{multicol}
\usepackage{algorithm}
\usepackage{algpseudocode}
\usepackage{soul}
\usepackage{array}
\usepackage{arydshln}
\usepackage{bbm}
\usetikzlibrary{bayesnet}
\usepackage{xcolor}
\usepackage{colortbl}
\usepackage[edges]{forest}
\usepackage{xcolor}
\usepackage[utf8]{inputenc}
\usepackage{tcolorbox}
\usepackage{xcolor}
\usepackage{pifont}
\usepackage{hyperref}
\usepackage{fontawesome5}
\usepackage{graphicx} 
\usepackage{svg}

\title{GeoMathCode: Understanding Interleaved Math-Code \\ Reasoning for Geometry Problem Solving}

\author{
Yingji Zhang$^{2,3,\dagger}$, Yong Dai$^{2}$, Andr\'{e} Freitas$^{1,3,4}$ \\
$^{1}$ Idiap Research Institute, Switzerland, $^{2}$ X-Humanoid, China \\
$^{3}$ Department of Computer Science, University of Manchester, UK\\
$^{4}$ Cancer Biomarker Centre, CRUK Manchester Institute, UK \\
\texttt{$\dagger$\{firstname.lastname\}3215@gmail.com}
}
\begin{document}

\maketitle

\begin{abstract}
Mathematical reasoning is a hallmark of human intelligence, requiring logical deduction, symbolic manipulation, and abstract thinking. Recent multimodal large language models (MLLMs) have demonstrated strong performance on geometry problems through multi-step reasoning. To better emulate human problem-solving, intermediate steps can incorporate auxiliary visual constructions, such as additional lines or points, which improve geometric interpretation and educational clarity. In this work, we introduce the GeoMathCode, where programmatic representations serve as intermediate visual outputs. We further conduct an in-depth analysis of the underlying reasoning geometry. Experimental results show that reasoning and code generation steps can be disentangled in the latent space, while supervised fine-tuning (SFT) makes the reasoning manifold more structured and informative. Moreover, hierarchical syntactic code structures emerge as disentangled latent subspaces, and contain more mathematical symbolic information than visual representations.
\end{abstract}
\section{Introduction}
Mathematical reasoning is widely regarded as a pinnacle of human intelligence, requiring a sophisticated interplay of logical deduction, symbolic manipulation, and abstract thinking. Recent advances in MLLMs have shown strong potential for mathematical reasoning \cite{guo2025deepseek}. Given a geometry problem, MLLMs can generate multi-step solutions. To further mirror human reasoning, each step can incorporate auxiliary visual constructions, such as additional lines or points, to provide clearer geometric interpretations and enhance educational effectiveness \cite{shi2025mathcanvas}.

In this scenario, traditional approaches to interleaved multimodal mathematical reasoning, such as BAGEL-Canvas \cite{shi2025mathcanvas}, often rely on pixel-based image representations for auxiliary visual generation. However, this fine-grained cross-modal generation paradigm is inherently difficult to align and verify \cite{chen2023lionempoweringmultimodal} and increases system complexity by introducing additional vision-specific generator \cite{deng2025emerging}, increasing the potential of error propagation and inconsistency across modalities.

To address these limitations, we investigate interleaved math-code reasoning, where executable code serves as an intermediate reasoning representation for geometry problem solving, as code-based representations are inherently precise and readily verifiable. However, in this task, an important question remains unclear: \textit{Does programmatic diagram generation fundamentally reshape the latent geometry of multimodal reasoning, or does it merely function as an auxiliary output representation?}

In particular, we seek to understand how interleaved math-code reasoning trajectories evolve across layers and reasoning steps, and whether symbolic code representations capture geometric concepts differently from visual representations.

\begin{figure*}[ht!]
  \includegraphics[width=\linewidth]{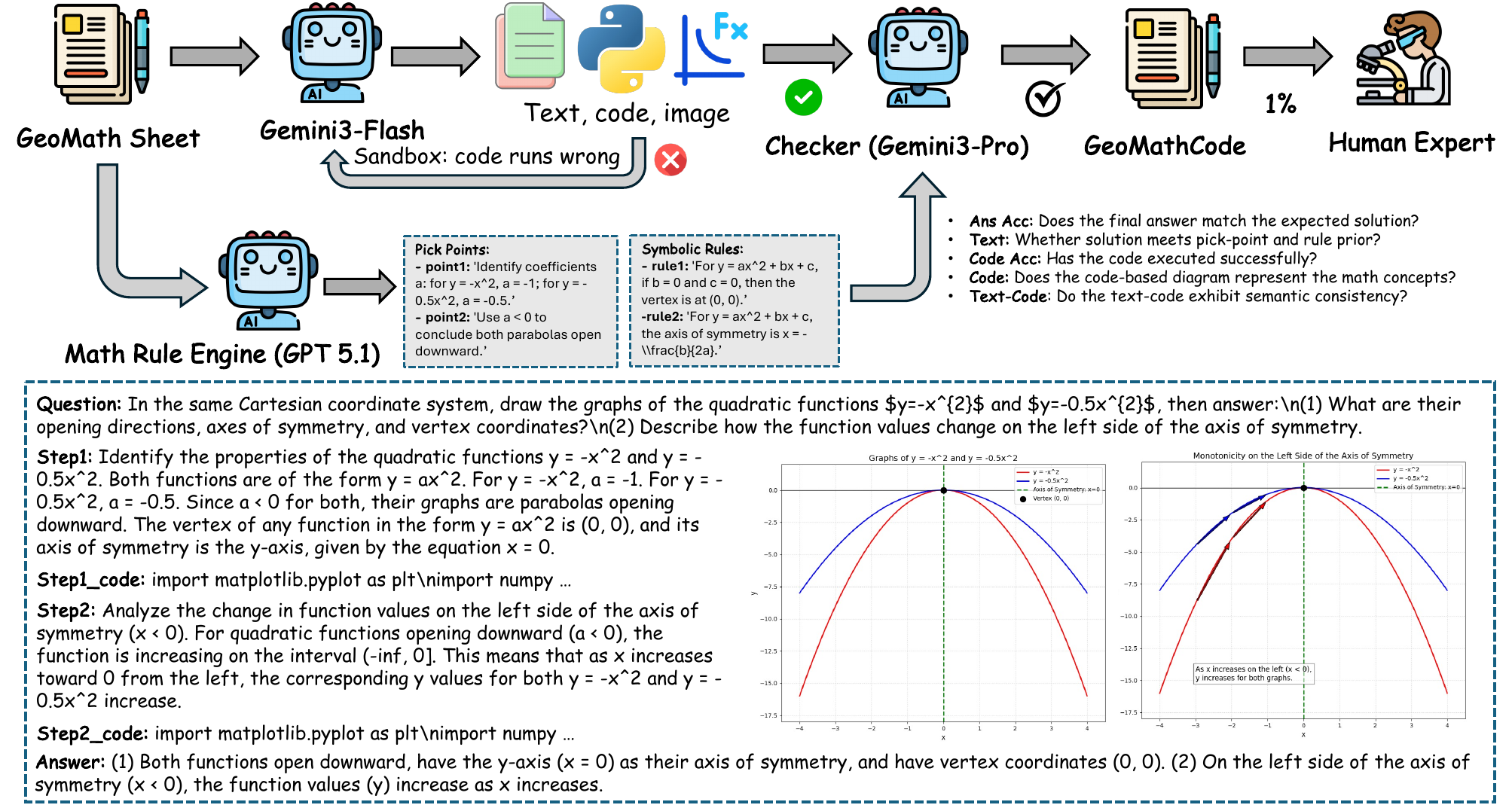}
  \caption{Pipeline overview. We construct \textbf{GeoMathCode} dataset for supervised fine-tuning.}
  \label{fig:pipeline}
\end{figure*}

To study these questions, we introduce a systematic pipeline for constructing a public multimodal math–code reasoning dataset, \textbf{GeoMathCode}, incorporating automated solution generation, soft rule-based evaluation, and code verification mechanisms. As illustrated in Figure~\ref{fig:pipeline}, given a GeoMath question, we employ the representative model Gemini~3-Flash to generate interleaved math--code reasoning solutions, while GPT-5.1 is used to produce key checkpoints for textual reasoning evaluation as well as factual mathematical rules for automatic code verification. Building upon this dataset, we conduct a systematic analysis of the latent geometry underlying math-code reasoning in MLLMs from both step-wise and stage-wise perspectives,  our experiments reveal several consistent findings.

\paragraph{Reasoning-code disentanglement.} First, reasoning and code-generation steps form disentangled latent subspaces across transformer layers, suggesting functional separation between mathematical reasoning and programmatic execution.

\paragraph{Reasoning manifold regularisation.} Second, SFT regularises the latent reasoning manifold by increasing global information capacity while simultaneously reducing local geometric complexity.

\paragraph{Hierarchical programmatic space.} Third, hierarchical programmatic syntactic structures emerge as structured latent subspaces, and latent code representations better capture fine-grained mathematical symbolic concepts than visual representations, revealing a progressive abstraction process from lexical syntax to execution-level semantics.

Overall, this work introduces a new dataset for multimodal math-code reasoning and provides insights into the latent geometric structure of reasoning processes, motivating future research on mechanistic interpretability in LLMs.





\section{Related Work} \label{sec:related} 
In this section, we review three topics, \textit{multimodal GeoMath datasets}, \textit{language geometry}, and \textit{code reasoning}, to further illustrate the motivation underlying our work.


\paragraph{Multimodal GeoMath Datasets.}
Early benchmarks such as Geometry3K~\cite{lu2021intergpsinterpretablegeometryproblem} and ScienceQA~\cite{lu2022learnexplainmultimodalreasoning} established foundational tasks for visual mathematical reasoning, while later datasets including MMMU~\cite{yue2024mmmumassivemultidisciplinemultimodal}, MathVista~\cite{lu2024mathvistaevaluatingmathematicalreasoning}, MathVision~\cite{wang2024measuring}, and MathVerse~\cite{zhang2024mavismathematicalvisualinstruction} further advanced the reasoning capabilities of multimodal models. However, most existing benchmarks focus on static problem--solution pairs and lack step-wise reasoning processes. Although MathCanvas~\cite{shi2025mathcanvas} introduces interleaved visual--textual reasoning, its reliance on pixel-based visualisations complicates verification and increases system complexity. In contrast, we focus on math--code reasoning, where Python-generated geometric diagrams provide structured and verifiable visual demonstrations.

\paragraph{Language Geometry.}
The geometric structure of language representations has been extensively studied~\cite{li2020sentence,chang2022geometry,jiang2024uncovering,li2022emergent,geva2022transformer,nanda2023emergent,trager2023linear,merullo2023language,turner2023activation}. Prior work shows that semantic representations often exhibit approximately linear structures in latent space, a phenomenon observed across modalities including vision~\cite{huh2024platonic}, language~\cite{trager2023linear}, and audio~\cite{liu2025nexus}, and across model architectures such as Transformers~\cite{ushio2021bert,merullo2023language,hernandez2023linearity}, VAEs~\cite{zhang-etal-2024-learning,zhang-etal-2025-quasi,zhang2026learning,carvalho2022learning}, and diffusion models~\cite{jing2022subspace}. Theoretical perspectives such as NTK-based reasoning control~\cite{zhang2026guiding} and weight-space disentanglement~\cite{ortizjimenez2023taskarithmetictangentspace} further support structured linear geometry in neural representations. Recent studies have examined geometric properties of reasoning in LLMs, including step-wise trajectories~\cite{sun2026llmreasoningtrajectoriesstepspecific}, layer-wise inference dynamics~\cite{ma2026reasoningemergesconstrainedinference}, and stage-wise training effects~\cite{li2026tracing,hoogland2024loss}. Building on these insights, we investigate the latent geometric structure underlying interleaved math--code reasoning processes.

\paragraph{Code Reasoning.}
Previous studies have shown that incorporating code during pre-training can improve agentic behaviour \cite{chen2025reinforcement}, tool use \cite{hong2024metagpt} and reasoning~\cite{liu2024training,aryabumi2025code}, while other work reports contradictory findings on reasoning performance~\cite{zhao2026reallyimprovesmathematicalreasoning}. Other studies further explore logical reasoning through programmatic representations~\cite{zhao2025unveiling}, and \citet{liang-etal-2025-grammar} demonstrate that grammar-based representations help LLMs distinguish subtle code differences. 
This work suggests that code functions primarily as an auxiliary diagram output rather than an active reasoning modality. Our experiments show that intermediate code representations do not improve textual reasoning performance and reasoning and code steps can be disentangled in the latent space, suggesting a functional separation from textual reasoning in latent space.
\section{Data Construction}

First, we propose an automated, verification-driven pipeline for constructing a high-quality GeoMath reasoning dataset. The pipeline is designed to ensure both the correctness and diversity of the generated samples. An overview of the overall process is illustrated in Figure~\ref{fig:pipeline}.

To ensure comprehensive coverage of high-quality GeoMath problems, we utilise MathCanvas \cite{shi2025mathcanvas} as the primary resource, which encompasses eight topics: Algebra, Analytic Geometry, Calculus, Trigonometry, Plane Geometry, Solid Geometry, Statistics, and Transformational Geometry. For each category, we first evaluate multiple state-of-the-art MLLMs as baselines (Gemini3-Flash, Gemini3-Pro, GPT-5.1, GPT-5.2) for automatic solution generation. As shown in Table~\ref{tab:baseline}, Gemini3 outperforms GPT5.1; therefore, we select Gemini3-Flash as the solution generator. GeoMath problems are then sourced from the dataset, and Gemini3-Flash is employed to generate solutions that integrate textual reasoning, executable Python code, and corresponding visualisations. A sandbox environment is used to ensure code validity by filtering out erroneous executions.

To enhance solution quality, we employ a rule-based engine (GPT-5.1) to extract key reasoning steps and relevant mathematical rules (Table~\ref{tab:rules}) for each problem. Subsequently, an automated checker (Gemini3-Pro) performs multi-level verification, including:
(i) assessing final answer correctness by comparing the outputs with reference solutions from MathCanvas, (ii) evaluating textual reasoning correctness (see prompts in Table~\ref{tab:prompt_rule}), and (iii) confirming successful execution of code.

For code execution, samples that fail to execute successfully are fed back into the generator for iterative regeneration of the solution. Only samples that pass all verification stages are retained, while textual reasoning samples must achieve at least 3/5 correctness. The dataset contains multiple-choice, fill-in-the-blank, and short-answer questions, with around 1\% further reviewed by human experts. The final \textbf{GeoMathCode} dataset consists of 10k training samples and 1k test samples, balanced with a 1:1 ratio of questions with and without diagrams.

\begin{table*}[ht!]
\centering
\begin{tabular}{l l c c c c c c}
\toprule
MLLM & Opt. & Ans Acc & Text & Code Acc & Code & Text Code & Avg. \\
\midrule
Qwen3-VL-Ins-8B    & -    & 0.32 & 0.38 & 0.59 & 0.41 & 0.37 & 0.41 \\
Qwen3-VL-Ins-8B    & SFT  & 0.48 & 0.57 & \textbf{0.97} & 0.70 & 0.68 & 0.71 \\
Qwen3-VL-Think-8B  & SFT  & 0.44 & 0.50 & 0.86 & 0.69 & 0.59 & 0.62 \\
Qwen3-VL-Ins-8B    & LoRA & 0.41 & 0.51 & 0.85 & 0.68 & 0.57 & 0.60 \\
\midrule
Qwen3.5-4B         & SFT  & 0.42 & 0.51 & 0.93 & 0.73 & 0.60 & 0.64 \\
Qwen3.5-9B         & SFT  & \textbf{0.55} & \textbf{0.65} & \textbf{0.97} & \textbf{0.76} & \textbf{0.73} & \textbf{0.73} \\
Qwen3.5-9B         & LoRA & 0.54 & 0.64 & 0.94 & 0.80 & 0.70 & 0.72 \\
\midrule
LLaVA-NEXT-8B      & -    & 0.00 & 0.00 & 0.00 & 0.00 & 0.00 & 0.00 \\
LLaVA-NEXT-8B      & SFT  & 0.15 & 0.23 & 0.83 & 0.43 & 0.26 & 0.38 \\
InternVL3.5-8B     & -    & 0.29 & 0.33 & 0.53 & 0.57 & 0.47 & 0.44 \\
InternVL3.5-8B     & SFT  & 0.48 & 0.58 & 0.96 & 0.70 & 0.68 & 0.68 \\
\bottomrule
\end{tabular}
\caption{Comparison of the performance of open-source MLLMs on the \textbf{GeoMathCode} dataset, evaluated using Gemini3-Pro due to its higher Spearman correlation between \textit{Ans Acc} and \textit{Text} scores. More experimental details are provided in Appendix~\ref{app:res}.} \label{tab:sft}
\end{table*}

\begin{table*}[ht!]
\centering
\begin{tabular}{lcccccc}
\toprule
MLLM & \multicolumn{2}{c}{Ans Acc-Text} & \multicolumn{2}{c}{Ans Acc-Text Code} & \multicolumn{2}{c}{Ans Acc-Code} \\ \hline
\multicolumn{7}{c}{\textit{without rule instruction}} \\
Qwen3-VL-Ins-8B-SFT & 0.48 & 0.48  & 0.22 & 0.23  & 0.20 & 0.21\\
Qwen3.5-9B-SFT & 0.49 & 0.47  & 0.21 & 0.22  & 0.21 & 0.23 \\
Qwen3.5-9B-LoRA & 0.50 & 0.47  & 0.22 & 0.23  & 0.22 & 0.21\\
InternVL3.5-8B-SFT & 0.48 & 0.47  & 0.23 & 0.24  & 0.23 & 0.24 \\ \hline
\multicolumn{7}{c}{\textit{with rule instruction}} \\
Qwen3-VL-Ins-8B-SFT & \textbf{0.70} & \textbf{0.70}  & 0.23 & 0.23 & 0.20 & 0.21 \\
Qwen3.5-9B-SFT & \textbf{0.69} & \textbf{0.70} & 0.20 & 0.22 & 0.21 & 0.20 \\
Qwen3.5-9B-LoRA & \textbf{0.70} & \textbf{0.70} & 0.23 & 0.23 & 0.22 & 0.23 \\
InternVL3.5-8B-SFT & \textbf{0.71} & \textbf{0.70} & 0.23 & 0.24 & 0.22 & 0.23 \\
\bottomrule
\end{tabular}
\caption{Pearson (left) and Spearman (right) correlation between different metrics. Same observation when using GPT5.1 and DeepSeek-V4-Flash as LLM evaluators in Table \ref{tab:corr1} and \ref{tab:corr2}, respectively.} \label{tab:corr}
\end{table*}

\begin{table*}[ht!]
\centering
\begin{tabular}{llcccc}
\toprule
Evaluator~1 & Evaluator~2 & Ans Acc & Text & Code & Text Code \\ \hline 
Gemini3-Pro & GPT-5.1 & 0.96 & 0.69 & 0.57 & 0.50 \\  
Gemini3-Pro & DeepSeek-V4-Flash & \textbf{0.99} & \textbf{0.94} & \textbf{0.70} & \textbf{0.70} \\ 
GPT-5.1 & DeepSeek-V4-Flash & 0.96 & 0.68 & 0.58 & 0.54 \\
\bottomrule
\end{tabular}
\caption{Spearman correlation between different LLM evaluators on the prediction of Qwen3.5-9B-SFT.} \label{tab:llm_corr}
\end{table*}

\section{Empirical Analysis} \label{sec:empirical}
In this work, we conduct experiments from two complementary perspectives: (1) post-training analysis; and (2) latent representation analysis.
\subsection{Post Training Analysis}

\paragraph{Experiment Setup.} We evaluate representative models from the Qwen3/3.5~\cite{yang2025qwen3}, LLaVA-Next~\cite{li2024llava}, and InternVL-3.5~\cite{wang2025internvl3} families. We consider both full-parameter supervised fine-tuning (SFT) and parameter-efficient tuning via LoRA~\cite{hu2022lora}. All models are trained in the LlamaFactory framework~\cite{zheng2024llamafactory} using a unified hyperparameter setup on 8 NVIDIA A800 GPUs. More details are provided in the Appendix \ref{app:exp}.

\paragraph{Evaluation Metrics.} We evaluate reasoning capabilities using five metrics: (1) \textit{Ans Acc}: final answer accuracy; (2) \textit{Text}: correctness of the textual reasoning, measured as the average of the point selection score and rule application score; (3) \textit{Code Acc}: accuracy of the executable Python code; (4) \textit{Code}: mathematical correctness of the generated code; and (5) \textit{Text--Code}: consistency between the textual reasoning step and the corresponding code.

For \textit{Ans Acc}, a mixed evaluation strategy is adopted: multiple-choice and numerical fill-in-the-blank questions are evaluated using exact match, while all other question types are assessed by MLLM evaluators. Metrics 1, 2, 4, and 5 are then computed using the scores from Gemini3-Pro as we find that the Gemini3-Pro has the higher Spearman correlation between \textit{Ans Acc} and \textit{Text} than GPT-5.1 and DeepSeek-V4-Flash. Finally, all scores are normalised to the range (0, 1). More details are provided in Appendix~\ref{app:res}.
\begin{figure*}[ht!]
  \includegraphics[width=\linewidth]{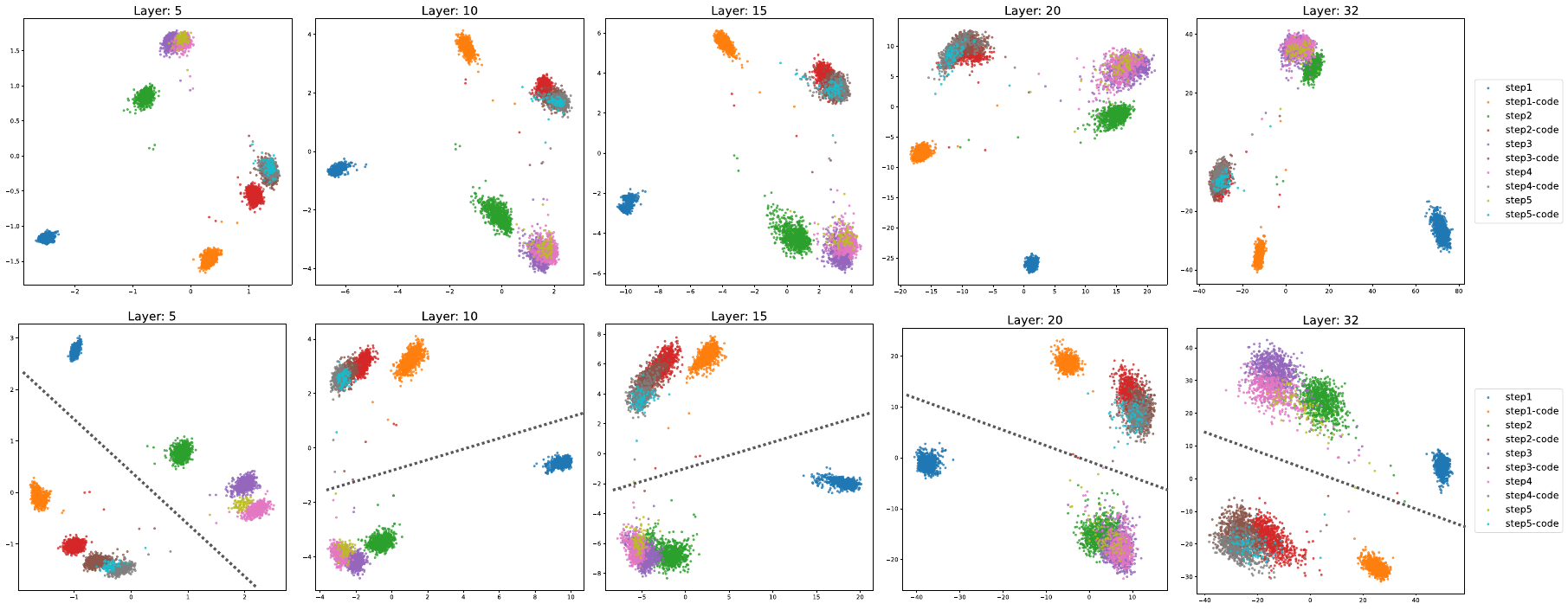}
  \caption{PCA visualisation of the reasoning step and code step geometry (Top: Qwen3.5-9B, bottom: Qwen3.5-9B-SFT). We can observe that reasoning and code steps can be disentangled in the latent space.}
  \label{fig:pca}
\end{figure*}

As shown in Table~\ref{tab:sft}, after the SFT stage, Qwen3.5 consistently outperforms all baseline models. Among them, Qwen3.5-9B achieves the best overall performance, followed by the LoRA-based setup, demonstrating the strong capability of the base model on this downstream task.

\paragraph{\textit{(1)~Point- and rule-based instructions during evaluation improve the Text-Ans Acc alignment.}} 
Furthermore, we examine the relationships among evaluation metrics by computing both Pearson and Spearman correlation coefficients. As reported in Table \ref{tab:corr}, incorporating pick-point-based instructions during evaluation improves the alignment between textual reasoning quality and final answer accuracy (Ans-Text). This finding suggests that such constraints enable the LLM evaluator to produce more consistent judgments, indicating a positive correlation between the logical coherence of reasoning process and the correctness of the final generated answer. More importantly, this observation highlights the potential of using intermediate pick-point and rule-based supervision signals for multi-step reinforcement learning.

Conversely, this incorporation does not improve the alignment between final-answer accuracy and text-code or Code, suggesting that intermediate code representations contribute less directly to final-answer prediction. This further indicates a potential functional disentanglement between reasoning and coding processes in the latent space, motivating our subsequent investigation of latent geometry.
\subsection{Reasoning Geometry Analysis}

\subsubsection{Reasoning Trajectory}

First, we analyse the overall geometry of multi-step reasoning in the latent space of MLLM.

\paragraph{Experiment Setup.} Prior studies have shown that reasoning-step markers contain informative geometric structures associated with reasoning trajectories \cite{sun2026llmreasoningtrajectoriesstepspecific}. Therefore, for each question in the test set, we generate the multi-step reasoning solution via greedy search and extract hidden representations of the token corresponding to the \textit{step} and \textit{step\_code} marker, denoted as $\mathbf{h}_t^{(\ell)}{}_{(\text{step}k)}$ and $\mathbf{h}_t^{(\ell)}{}_{(\text{step}k\_\text{code})}$, which captures the model’s initial information transitioning to the reasoning step. The ordered sequence $\mathbf{h}_t^{(\ell)}{}_{(\text{step}1)}$, $\mathbf{h}_t^{(\ell)}{}_{(\text{step}1\_\text{code})}$, $\dots$, $ \mathbf{h}_t^{(\ell)}{}_{(\text{stepN})}$, and $ \mathbf{h}_t^{(\ell)}{}_{(\text{stepN}\_\text{code})}$ thus provides a series of reasoning trajectory of the model’s internal representations.
\begin{figure*}[ht!]
  \includegraphics[width=\linewidth]{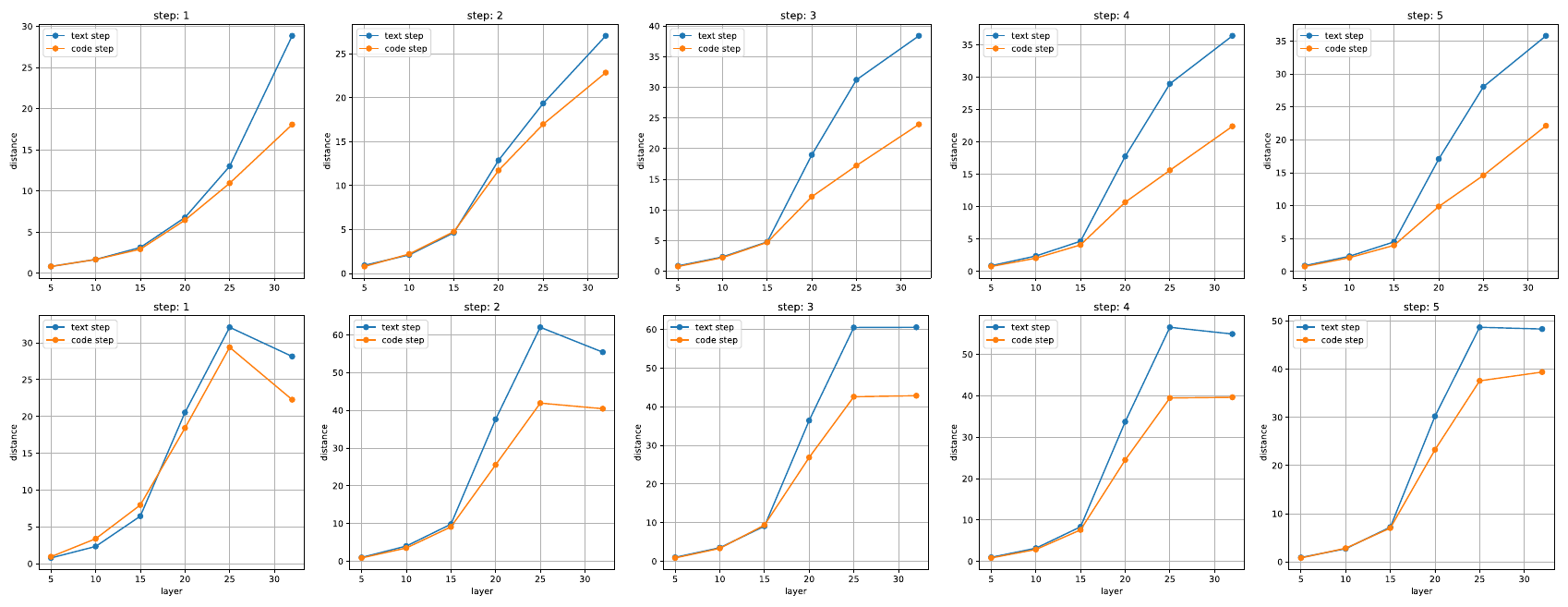}
  \caption{Average euclidean distance of different steps at different layers (Top: Qwen3.5-9B, bottom: Qwen3.5-9B-SFT). Same observation for Qwen3-VL-8B and InternVL3.5-8B in Figure \ref{fig:dist_qwen3} and \ref{fig:dist_internvl}.}
  \label{fig:dist}
\end{figure*}
\begin{figure*}[ht!]
  \includegraphics[width=\linewidth]{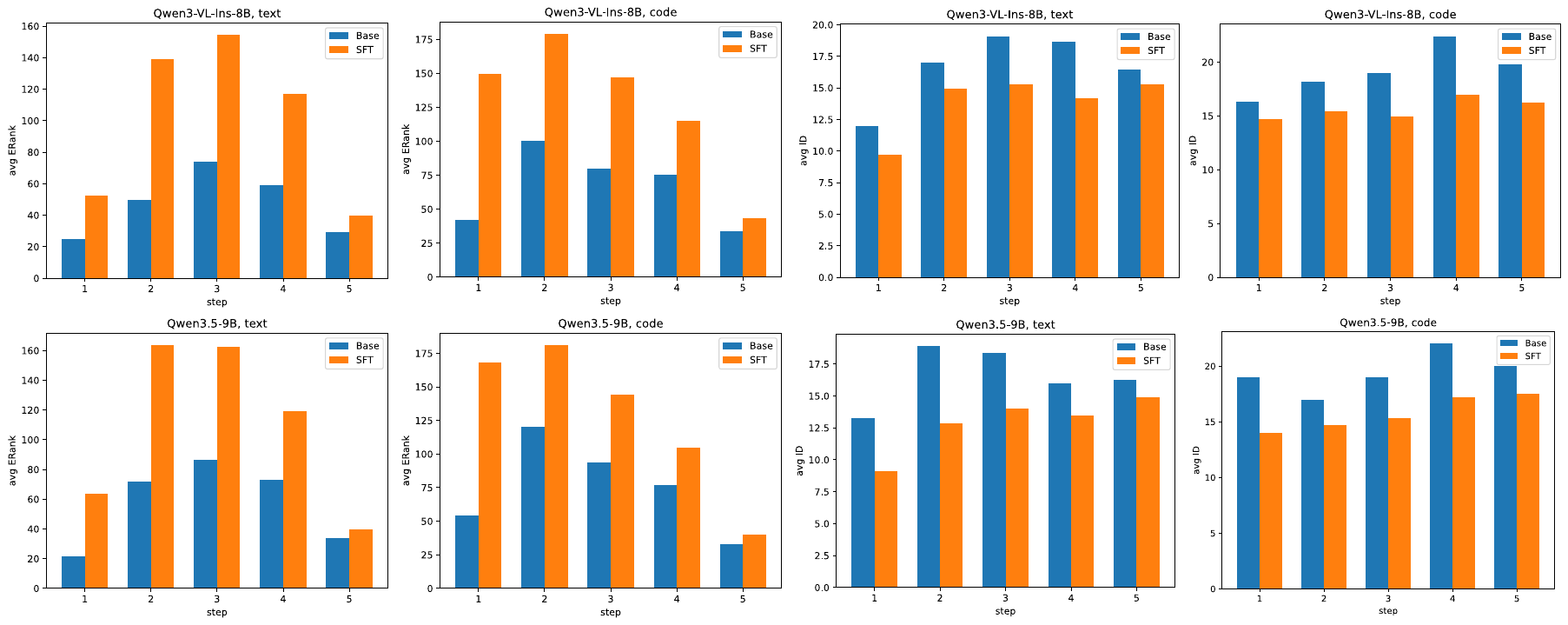}
  \caption{ERank $\uparrow$ (left) and ID $\downarrow$ (right) for Qwen3-VL-Ins-8B (top) and Qwen3.5-9B (bottom).}
  \label{fig:inf}
\end{figure*}
\paragraph{\textit{\textbf{(1) Reasoning and code steps are geometrically separable in the latent space.}}} Figure~\ref{fig:pca} shows clear separability across reasoning stages and between reasoning and code representations. Notably, Step~1 forms a geometrically isolated region across layers, suggesting its role in semantic grounding and problem initialisation, while later reasoning steps progressively converge into a shared reasoning manifold.

\begin{table}[ht!]
\centering
\resizebox{7.7cm}{!}{
\begin{tabular}{lccc}
\toprule
MLLM & Ans Acc & Text & Avg. \\ \hline
\multicolumn{4}{c}{\textit{with code}} \\
Qwen3-VL-Ins-8B & 0.32 & 0.38 &  0.35 \\
Qwen3-VL-Ins-8B-SFT & 0.48 & 0.57 & 0.53 \\
Qwen3.5-9B-SFT & 0.55 & 0.65 & 0.60 \\ \hline
\multicolumn{4}{c}{\textit{without code}} \\
Qwen3-VL-Ins-8B & 0.33 & 0.38 & 0.36 \\
Qwen3-VL-Ins-8B-SFT & 0.48 & 0.57 & 0.53 \\
Qwen3.5-9B-SFT & 0.54 & 0.65 &  0.60 \\
\bottomrule
\end{tabular}
}
\caption{Performance of GeoMathCode without code generation.} \label{tab:code_ablation}
\end{table}


In addition to PCA, we also compared with Qwen3 and 3.5 with or without code generation setup to further evaluate the separation between text and code reasoning. As shown in Table \ref{tab:code_ablation}, both settings achieve highly similar performance, suggesting a functional separation between textual reasoning and diagram generation. Specifically, the model primarily relies on textual reasoning trajectories for final-answer prediction, while intermediate code generation mainly serves as an auxiliary executable representation rather than an active reasoning modality. This behaviour differs from visual interleaved chain-of-thought settings, where visual modalities can directly contribute to and guide the reasoning process \cite{hu2024visual,deng2025emerging,liu2025thinking}. Instead, these findings align with recent studies showing that programmatic data do not necessarily provide additive improvements to model capabilities in complex math reasoning task \cite{zhao2026reallyimprovesmathematicalreasoning}.

\paragraph{\textit{\textbf{(2) Deep layers exhibit a more diverse reasoning regime.}}} To better understand the reasoning step geometry, we measure the average euclidean distance between central point and others within each reasoning step cluster. As illustrated in Figure \ref{fig:dist}, we observe that the intra-step representation distance progressively increases with layer depth, suggesting that the model enters a more diverse and computation-dependent reasoning regime during later layers of problem solving. The comparatively smaller distances observed in code reasoning further suggest that program-based reasoning follows a more constrained and structurally regular latent trajectory than free-form textual reasoning.

\begin{figure*}[ht!]
  \includegraphics[width=\linewidth]{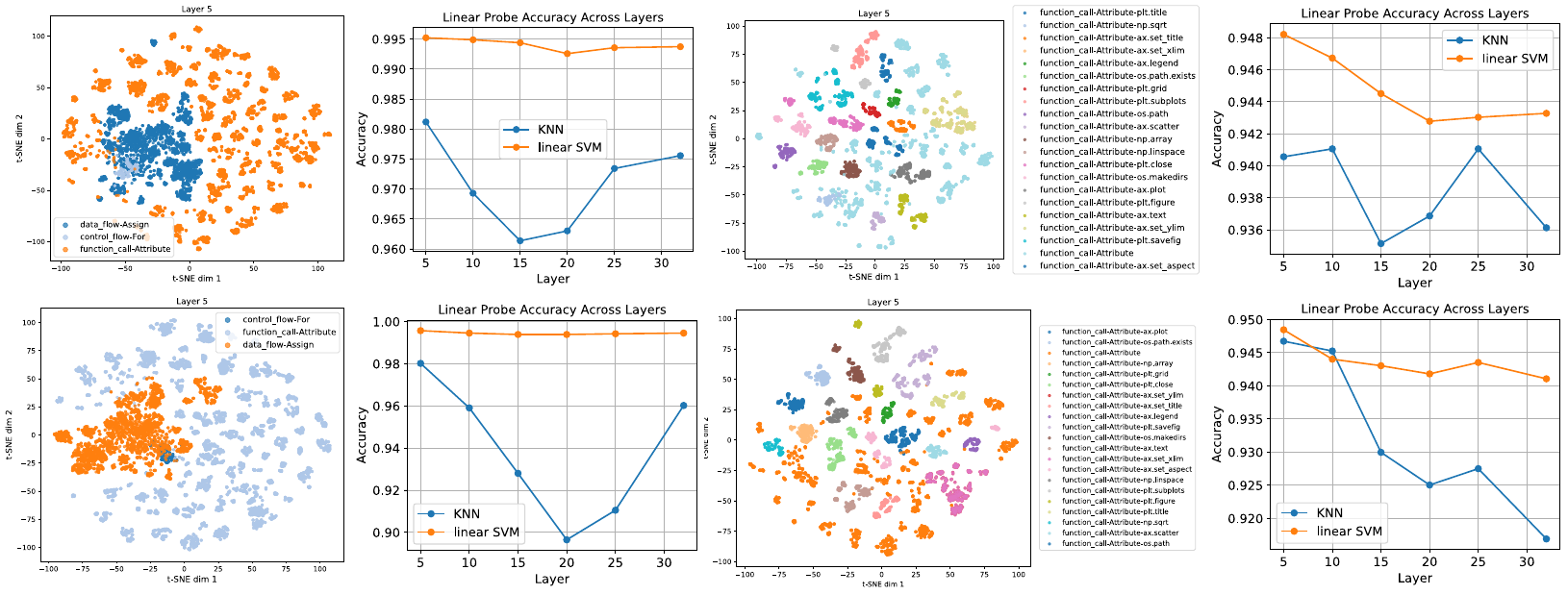}
  \caption{Code syntax space visualisation and linear probing results for Qwen3.5-9B (Top) and Qwen3.5-9B-SFT (Bottom). Left: visualisation of the representation spaces corresponding to high-level program semantics: \texttt{control\_flow}, \texttt{function\_call}, and \texttt{data\_flow}; right: fine-grained analysis of the \texttt{function\_call} category.}
  \label{fig:syntax}
\end{figure*}
\paragraph{\textit{\textbf{(3) SFT makes reasoning steps manifold more informative and regular.}}} Interestingly, we can observe that SFT stage simultaneously increases the intra-step representation distance. This suggests that SFT does not merely compress reasoning representations into more compact manifolds, but instead promotes richer intermediate reasoning information. To quantify the information capacity and geometric complexity of reasoning representations, we measure both the \textit{Effective Rank (ERank)} and the \textit{Intrinsic Dimensionality (ID)} of hidden states across layers and reasoning steps. Given the covariance matrix $\mathbf{\Sigma}$ of hidden representations, let $\{\lambda_i\}_{i=1}^{d}$ denote its eigenvalue spectrum and define the normalised spectrum as
$
p_i = \frac{\lambda_i}{\sum_{j=1}^{d}\lambda_j}.
$
The \textit{Effective Rank (ERank)} is defined as the exponential of the spectral entropy:
$
\mathrm{ERankank}(\mathbf{\Sigma})
=
\exp\left(
-\sum_{i=1}^{d} p_i \log p_i
\right)
$
, which characterises how evenly information spreads across eigen-directions. A larger ERank indicates that the representation manifold spreads across more informative directions, reflecting richer and more diverse latent patterns.
For the \textit{Intrinsic Dimensionality (ID)}, it is defined as
$
\mathrm{ID}(\mathbf{\Sigma})
=
\frac{\left(\sum_{i=1}^{d}\lambda_i\right)^2}
{\sum_{i=1}^{d}\lambda_i^2}
$
It measures the number of dominant dimensions contributing to the representation space (i.e., local manifold complexity). Higher ID corresponds to more complex local geometry, while lower ID indicates more regular and constrained representation structures.

As shown in Figure \ref{fig:inf}, SFT increases the ERank of reasoning representations while reducing their ID, suggesting a more informative yet geometrically regular reasoning manifold. This suggests that SFT does not simply increase representational complexity in an unconstrained manner, but instead promotes a more structured organisation of the latent reasoning manifold, reflecting increasingly constrained and canonical reasoning trajectories after fine-tuning.

This observation provides several important insights. First, stronger reasoning representations are not necessarily associated with higher geometric complexity. Instead, effective reasoning appears to require a balance between information diversity and geometric regularity. Second, rather than compressing representations into smaller latent regions, SFT reorganises reasoning trajectories into more canonical and reusable computational paths, which provides a strong motivation for future study in intermediate supervision in reasoning training.

\subsubsection{Geometrical Structure of Code Space}
Next, we focus on the geometric structure of code space, as it follows structured and deterministic processes. Studying code space therefore provides clearer insights into the syntactic and semantic structures encoded in latent representations.

\paragraph{Experiment Setup.} In the experiment, we extract the generated code at the middle reasoning step and employ Abstract Syntax Tree (AST) parser to automatically identify the structure information (e.g., function call, data structure, etc.) and extract the corresponding hidden representations.

First, we analyse high-level syntax categories, including control flow (e.g., \texttt{for} loop), function calls (e.g., \texttt{np.array}), and data flow (e.g., variable assignments), as well as the fine-grained structure within the \textit{function call} category. We use KNN and linear SVM to evaluate the local structure and global linear separability, respectively.

\paragraph{\textit{(1)~Hierarchical syntactic code structures can be disentangled and represented as latent subspaces.}}
As illustrated in Figure~\ref{fig:syntax} (t-SNE visualisation), distinct fine-grained programmatic syntactic structures (e.g., \texttt{function\_call}) can be separated in the latent space, suggesting that the model organises different code semantics into structured representation subspaces. Meanwhile, certain high-level structures display partial overlap due to their compositional nature. For example, \texttt{for} loops are not purely control-flow operations, but also implicitly involve variable assignment, iteration-state updates, and function-call semantics. Consequently, their representations partially overlap with both \texttt{data\_flow} and \texttt{function\_call} subspaces, indicating that intermediate code representations encode compositional execution semantics rather than isolated syntactic categories alone.

\paragraph{\textit{(2)~Latent code representations progressively lose locally compact lexical clustering while preserving globally linearly semantic structures.}} Next, as shown in Figure \ref{fig:syntax} (accuracy curve), we can observe that SVM has a higher accuracy than KNN, a higher linear SVM accuracy indicates that information remains globally linear in the latent space. In other words, the model has not completely discarded specific information, e.g., different function calls can still be separated by global semantic directions. In contrast, the decline in KNN accuracy suggests that local neighborhood structures become increasingly mixed in deeper layers. Function calls with similar computational roles, such as plotting- or rendering-related APIs, begin to overlap within local regions of the latent space. This indicates that the model no longer represents APIs as isolated lexical categories, but instead reorganises them into shared execution-semantic manifolds based on their functional roles in program execution.

The results reveal a potential hierarchical abstraction dynamic in code representations. High-level semantic abstractions, such as control-flow, data-flow, and function-call categories, remain relatively preserved throughout layer evolution, while fine-grained lexical and syntactic information becomes dynamically entangled or separated across layers. 

These findings may inspire future studies on the mechanisms of function-calling and executable reasoning in LLMs, where latent code-generation states may resemble structured state-transition processes similar to state machines. Mechanistic interpretability approaches, such as sparse autoencoders and circuit analysis, may further help uncover the internal computational structures underlying these executable reasoning behaviours.

\paragraph{\textit{(3)~Latent code representations better capture fine-grained mathematical symbolic concepts than visual representations.}} Finally, we evaluate whether code representations preserve more fine-grained mathematical symbolic information than visual representations. For each mathematical question, we assign labels based on the exact occurrence of specific symbolic categories, including \texttt{frac}, \texttt{sqrt}, \texttt{circ}, \texttt{triangle}, \texttt{angle}, \texttt{begin{cases}}, \texttt{sin}, and \texttt{overrightarrow}, within mathematical formulas delimited by ($\$\texttt{formula}\$$). Each symbolic corresponds to distinct metathetical concept.

We then extract the corresponding pooled code representations and rendered visual representations by feeding them into an MLLM encoder. A classifier is subsequently trained to evaluate the representational capabilities of each modality. As reported in Table~\ref{tab:math_symbolic}, code representations achieve higher accuracy, indicating that latent code embeddings capture fine-grained mathematical symbolic concepts more effectively than visual representations. Furthermore, compared to the base model, SFT enhances the model's ability to represent fine-grained mathematical expressions.
\begin{table}[ht!]
\centering
\small
\begin{tabular}{lcccc}
\toprule
classifier & Code & Image & Code & Image  \\ \hline
\multicolumn{5}{c}{\textit{Layer 10}} \\
KNN & 0.66 & 0.65 & \textbf{0.67} & 0.65 \\
SVM & 0.82 & 0.77 & \textbf{0.83} & 0.77 \\
RF & 0.69 & 0.67 & \textbf{0.71} & 0.66 \\ \hline
\multicolumn{5}{c}{\textit{Layer 20}} \\
KNN & 0.66 & 0.63 & \textbf{0.70} & 0.67 \\
SVM &0.81 & 0.76 & \textbf{0.83} & 0.78 \\
RF & 0.70 &0.66 & \textbf{0.72} & 0.66 \\ \hline
\multicolumn{5}{c}{\textit{Layer 32}} \\
KNN & \textbf{0.66} & 0.65 & \textbf{0.66} & 0.65 \\
SVM & 0.83 & 0.75 & \textbf{0.84} & 0.75 \\
RF & 0.70 & 0.69 & \textbf{0.71} & 0.65 \\
\bottomrule
\end{tabular}
\caption{Fine-grained probing on math symbolic concepts (left: Qwen3.5-9B, right: Qwen3.5-9B-SFT), where RF is random forest.} \label{tab:math_symbolic}
\end{table}



\section{Conclusion}
In this work, we introduce the GeoMathCode dataset, where programmatic representations serve as intermediate visual outputs. We further conduct an in-depth analysis of the underlying reasoning process. Experimental results show that reasoning and code generation steps can be disentangled in the latent space, while SFT makes the reasoning manifold more structured and informative. Moreover, hierarchical syntactic code structures emerge as disentangled latent subspaces, suggesting a hierarchical abstraction process during reasoning. These findings can inspire our future studies on the mechanistic interpretability of code execution, pick-point-based reinforcement learning, etc.

\section*{Limitations}
This work has several limitations. First, the automatic evaluation pipeline relies heavily on MLLM-based judgment, which may introduce biases and inconsistencies in assessing reasoning quality. Although point- and rule-based instructions improve evaluation alignment, the reliability and robustness of LLM evaluators remain limited compared to human expert annotation. In particular, the evaluator may fail to accurately distinguish subtle logical errors or overestimate superficially coherent reasoning traces. Second, the overall final answer accuracy remains relatively low. This suggests that improvements in reasoning capability and final answer prediction are still necessary. To address these limitations, future work will focus on developing automatic symbolic evaluation methods for multi-step reasoning processes and exploring reinforcement learning approaches, such as GRPO \cite{shao2024deepseekmath} and DAPO \cite{yu2026dapo}, to further improve final answer accuracy and reasoning reliability.

\bibliography{acl}
\bibliographystyle{acl_natbib}

\appendix
\section{Appendix}
\subsection{Experiment Details} \label{app:exp}

\paragraph{GeoMathCode Dataset.} After tokenisation and applying the chat template, the average sequence length of GeoMathCode is 3020 tokens. In terms of reasoning depth (measured on the training set), 9683 samples has three steps, 5717 samples contain four reasoning steps, 946 samples contain five reasoning steps, and 67 samples contain six reasoning steps, indicating that the dataset predominantly consists of long-form multi-step reasoning trajectories. Table \ref{tab:pick} and Table \ref{tab:rules} present representative examples of extracted pick-points and symbolic mathematical rules used in the automatic verification pipeline.
\begin{table}[ht!]
\centering
\begin{tcolorbox}[fontupper=\small, fontlower=\scriptsize]
\textbf{Atomic Rules:}

\textbf{1.} $\text{Perimeter}(\triangle ABC) = AB + BC + CA.$ \\
\textbf{2.} $r = d / 2$ \\
\textbf{3.} A linear pair of adjacent angles forms a straight angle: $m\angle 1 + m\angle 2 = 180^\circ$. \\

\textbf{Conditional Rules:} 

\textbf{1.}~Definition of midpoint: If $M$ is midpoint of segment $PQ$, then $PM = MQ$ and $M$ lies on $PQ$. \\
\textbf{2.}~All radii of a circle are equal: if $O$ is the center, then $OA = OB = OC = r.$ \\
\textbf{3.} If all three sides of a triangle are equal, then the triangle is equilateral.
\end{tcolorbox}
\caption{Example of math rules for plane geometry category.}
\label{tab:rules}
\end{table}

\begin{table}[ht!]
\centering
\begin{tcolorbox}[fontupper=\small, fontlower=\scriptsize]
\textbf{1.} Recognizes both functions are of the form $y = ax^2$ with a = -1 and a = -0.5. \\
\textbf{2.} Uses $a < 0$ to conclude both parabolas open downward. \\
\textbf{3.} Identifies vertex at (0, 0) for functions of the form $y = ax^2$ with no linear or constant term. \\
\textbf{4.} Identifies axis of symmetry as x = 0 (the y-axis). \\
\textbf{5.} States that for a downward-opening parabola, the function is increasing on $(-\infty, 0]$. \\
\textbf{6.} Describes that as x increases toward 0 from the left (x < 0), the function values y increase for both functions.
\end{tcolorbox}
\caption{Example of math pick points.}
\label{tab:pick}
\end{table}

\paragraph{Training Setups.}
For full-parameter SFT, model is trained using DeepSpeed ZeRO-3 optimisation with a per-device batch size of 1 and gradient accumulation over 8 steps. For LoRA-based training, LoRA adapters are applied to all trainable modules of model, with a per-device batch size of 4 and gradient accumulation over 2 steps. Following the LoRA+ setting, we use a learning-rate ratio of 16.0 for LoRA parameters. For both settings, we employ the AdamW optimiser with a cosine learning-rate scheduler, an initial learning rate of $5\times10^{-5}$, a warmup ratio of 0.1, and gradient clipping with a maximum gradient norm of 1.0. All models are trained for 2 epochs using bfloat16 precision. To support long interleaved math-code reasoning trajectories, the maximum sequence length is set to 100k tokens. All experiments are conducted on 8 NVIDIA A800-SXM4-80GB GPUs. For the classifier, we use sklearn to implement them using default hyper-parameters. 

\paragraph{Prompts.} 
The prompts used for evaluating textual reasoning correctness (Text), code correctness (Code), text-code consistency (Text-Code), final answer accuracy (Ans Acc), and rule extraction are provided in Table \ref{tab:prompt_text}, Table \ref{tab:prompt_code}, Table \ref{tab:prompt_ans}, and Table \ref{tab:prompt_rule}, respectively.

\begin{table*}[ht!]
\centering
\begin{tcolorbox}[fontupper=\small, fontlower=\scriptsize]
You are a geometry auto-marker. \\

Input: \\
* problem \\
* student\_solution \\
* reference\_dict= \\
{"overall\_strategy", "key\_pick\_points", "rules"} \\

Task: \\
Mark the student solution against the reference\_dict. \\

Rules: \\
* Accept equivalent valid reasoning \\
* Accept different order of steps \\
* Accept implicit theorem use if clearly applied \\
* Do not require theorem names \\
* Do not award full credit for unsupported conclusions \\
* Penalize theorem misuse, missing essential checkpoints, and logical gaps \\
* Base marks mainly on key\_pick\_points \\

Scoring rules: \\
- 5 = fully correct \\
- 4 = minor issue, overall mathematically correct \\
- 3 = partially correct, noticeable error \\
- 2 = major error \\
- 1 = mostly incorrect \\
- 0 = completely incorrect or missing \\

Return STRICT JSON only: \\
- "overall\_strategy\_assessment": 0-5, \\
- "pick\_point\_assessment": 0-5, \\
- "rule\_assessment": 0-5,
\end{tcolorbox}
\caption{
Example of the prompt used for textual reasoning evaluation. The final Text score is computed as the average of the \texttt{pick\_point\_assessment} and \texttt{rule\_assessment} scores.}
\label{tab:prompt_text}
\end{table*}

\begin{table*}[ht!]
\centering
\begin{tabular}{lccccccc}
\toprule
\textbf{MLLM} & \textbf{Opt.} & \textbf{Ans Acc} & \textbf{Text} & \textbf{Code Acc} & \textbf{Code} & \textbf{Text Code} & \textbf{Avg.} \\
\midrule
Qwen3-VL-Ins-8B & - & 0.33 & 0.38 & 0.59 & 0.47 & 0.41 & 0.44 \\
Qwen3-VL-Ins-8B & SFT & 0.49 & 0.60 & \textbf{\textcolor{black}{0.97}} & 0.80 & 0.67 & 0.71 \\
Qwen3-VL-Think-8B & SFT & 0.45 & 0.52 & 0.87 & 0.72 & 0.61 & 0.63 \\
Qwen3-VL-Ins-8B & LoRA & 0.42 & 0.53 & 0.86 & 0.70 & 0.58 & 0.62 \\ \hline
Qwen3.5-4B & SFT & 0.44 & 0.53 & 0.94 & 0.75 & 0.62 & 0.66 \\
Qwen3.5-9B & SFT & 0.56 & \textbf{\textcolor{black}{0.65}} & \textbf{\textcolor{black}{0.97}} & \textbf{\textcolor{black}{0.84}} & \textbf{\textcolor{black}{0.73}} & \textbf{\textcolor{black}{0.75}} \\
Qwen3.5-9B & LoRA & \textbf{\textcolor{black}{0.57}} & \textbf{\textcolor{black}{0.65}} & 0.94 & 0.82 & 0.72 & 0.74 \\ \hline
LLaVA-NEXT-8B & - & 0.00 & 0.00 & 0.00 & 0.00 & 0.00 & 0.00 \\
LLaVA-NEXT-8B & SFT & 0.16 & 0.25 & 0.85 & 0.45 & 0.28 & 0.40 \\
InternVL3.5-8B & - & 0.30 & 0.35 & 0.54 & 0.60 & 0.50 & 0.46 \\
InternVL3.5-8B & SFT & 0.49 & 0.61 & 0.96 & 0.79 & 0.67 & 0.70 \\
\bottomrule
\end{tabular}
\caption{GPT-5.1 evaluation results.} \label{tab:gpt5.1}
\end{table*}


\begin{table*}[ht!]
\centering
\begin{tabular}{l l c c c c c c}
\toprule
\textbf{MLLM} & \textbf{Opt.} & \textbf{Ans Acc} & \textbf{Text} & \textbf{Code Acc} & \textbf{Code} & \textbf{Text Code} & \textbf{Avg.} \\
\midrule
Qwen3-VL-Ins-8B    & -    & 0.30 & 0.36 & 0.58 & 0.39 & 0.35 & 0.39 \\
Qwen3-VL-Ins-8B    & SFT  & 0.46 & 0.55 & 0.97 & 0.68 & 0.66 & 0.69 \\
Qwen3-VL-Think-8B  & SFT  & 0.42 & 0.48 & 0.85 & 0.66 & 0.57 & 0.60 \\
Qwen3-VL-Ins-8B    & LoRA & 0.39 & 0.49 & 0.84 & 0.65 & 0.55 & 0.58 \\
\midrule
Qwen3.5-4B         & SFT  & 0.40 & 0.49 & 0.92 & 0.71 & 0.58 & 0.62 \\
Qwen3.5-9B         & SFT  & 0.55 & 0.64 & 0.97 & 0.69 & 0.69 & 0.71 \\
Qwen3.5-9B         & LoRA & 0.52 & 0.62 & 0.93 & 0.77 & 0.67 & 0.70 \\
\midrule
LLaVA-NEXT-8B      & -    & 0.00 & 0.00 & 0.00 & 0.00 & 0.00 & 0.00 \\
LLaVA-NEXT-8B      & SFT  & 0.13 & 0.21 & 0.81 & 0.41 & 0.24 & 0.36 \\
InternVL3.5-8B     & -    & 0.27 & 0.31 & 0.51 & 0.54 & 0.44 & 0.41 \\
InternVL3.5-8B     & SFT  & 0.46 & 0.56 & 0.96 & 0.68 & 0.66 & 0.66 \\
\bottomrule
\end{tabular}
\caption{DeepSeek-V4-Flash evaluation results.}
\label{tab:deepseekv4flash}
\end{table*}

\begin{table*}[ht!]
\centering
\begin{tcolorbox}[fontupper=\small, fontlower=\scriptsize]
You are a Python code auto-marker for geometry/math visualization tasks. \\

Input: \\
- the GeoMath problem \\
- the multi-step solution \\
- the Python code used to generate the image for each step \\

Task: \\
Verify whether the code is mathematically correct. \\

Score each category from 0 to 5: \\
- equation: whether the implemented functions exactly match the problem equations \\
- properties: whether key properties derived from the formulas are correct, such as vertex, axis of symmetry, and opening direction, etc. \\
- points: whether sample points used or implied by the code agree with the true function values \\
- range: whether the plotting range is appropriate \\
- annotations: whether labels, arrows, symmetry lines, marked vertices, and other annotations are mathematically consistent with the plotted functions \\
- consistency: whether the textual reasoning step and the corresponding code are consistent with each other \\

Scoring rules: \\
- 5 = fully correct \\
- 4 = minor issue, overall mathematically correct \\
- 3 = partially correct, noticeable error \\
- 2 = major error \\
- 1 = mostly incorrect \\
- 0 = completely incorrect or missing \\

Return strict JSON with fields: \\
- "equation": 0-5, \\
- "properties": 0-5, \\
- "points": 0-5, \\
- "range": 0-5, \\
- "annotations": 0-5, \\
- "consistency": 0-5
\end{tcolorbox}
\caption{
Example of the prompt used for code correctness evaluation. The final Code score is computed as the average of the \texttt{equation}, \texttt{properties}, \texttt{points}, \texttt{range}, and \texttt{annotations} scores, while the Text-Code score corresponds to the \texttt{consistency} score.
}
\label{tab:prompt_code}
\end{table*}
\begin{table*}[ht!]
\centering
\begin{tcolorbox}[fontupper=\small, fontlower=\scriptsize]
You are a strict grading assistant. \\
Given a question, a ground truth answer, and a predicted answer, decide if the predicted answer is semantically correct.

Rules: \\
- Focus only on correctness relative to the ground truth. \\
- Ignore style, wording, and extra explanation. \\

Finally, you should output *correct* or *incorrect* ONLY. \\
Return strict JSON with fields: \\
- "ans\_match": correct/incorrect
\end{tcolorbox}
\caption{
Example of the prompt used for final answer evaluation. This prompt is only used for marking short-description question-answer pairs.
}
\label{tab:prompt_ans}
\end{table*}
\begin{table*}[ht!]
\centering
\begin{tcolorbox}[fontupper=\small, fontlower=\scriptsize]
Extract the reasoning structure and pick points from this geometry math solution. \\

Given: \\

A geometry problem \\
A multi-step solution \\

Produce: \\
- overall strategy \\
- key pick points used for marking the solution \\
- theorem/rule used at each step \\

Return strict JSON with fields: \\
- overall\_strategy: "...", \\
- key\_pick\_points: ["...", ...],  \\
- rules: [{step, rule, rule\_type, explicit\_or\_implicit}], \\

Additional constraints: \\
- Rules MUST be universal (not problem-specific wording) \\
- Rules MUST be symbolic whenever possible \\
- If symbolic form is not possible, use minimal canonical phrasing \\
- Do NOT include explanations, only abstractions of rules \\
\end{tcolorbox}
\caption{
Example of the prompt used for extracting symbolic mathematical rules and reasoning pick-points from multi-step geometry solutions.
}
\label{tab:prompt_rule}
\end{table*}

\subsection{Human Evaluation Protocol}

To validate the reliability of the LLM-based automatic scoring pipeline, we conduct a human evaluation on a randomly sampled subset of GeoMathCode. The goal is to assess whether automatic scores are consistent with expert judgments on textual reasoning correctness, code-generated diagram correctness, and text-code consistency.

\paragraph{Sample selection.}
We sample 100 examples from the test set, stratified by geometry topic and question type. Each example contains the original problem, the reference answer, the generated multi-step solution, the generated Python code for each reasoning step, and the rendered diagram. The automatic LLM scores are hidden from annotators during evaluation.

\paragraph{Annotators.}
Each sample is evaluated by one PhD-level annotator with mathematical or geometry problem-solving background. Annotator is instructed to focus on mathematical correctness rather than stylistic quality.

\paragraph{Evaluation dimensions.}
Annotator scores each sample along three dimensions:

\paragraph{\textbf{1.~Textual reasoning correctness.}}
Whether the generated reasoning follows valid mathematical logic, uses appropriate geometric facts, and reaches intermediate conclusions correctly.

\paragraph{\textbf{2.~Final answer correctness.}}
Whether the final answer is equivalent to the reference answer, allowing equivalent expressions, reordered items, or semantically identical explanations.

\paragraph{\textbf{3.~Diagram correctness.}}
Whether the Python diagram correctly include the mathematical objects, including equations, points, angles, curves, annotations, auxiliary constructions, and plotting ranges.

\paragraph{Scoring rubric.}
Each dimension is scored on a 0--5 scale:

\begin{itemize}
    \item \textbf{5}: Fully correct; no mathematical or consistency errors.
    \item \textbf{4}: Mostly correct; only minor omissions or harmless presentation issues.
    \item \textbf{3}: Partially correct; contains noticeable errors but preserves the main reasoning direction.
    \item \textbf{2}: Major errors; reasoning, code, or diagram is substantially flawed.
    \item \textbf{1}: Mostly incorrect; only minimal relevant content is correct.
    \item \textbf{0}: Missing, invalid, irrelevant, or completely incorrect.
\end{itemize}

\paragraph{Annotation guidelines.}
Annotator is asked to apply the following rules:

\begin{itemize}
    \item Accept mathematically equivalent reasoning even if it differs from the reference solution.
    \item Do not require theorem names if the theorem is correctly applied.
    \item Penalize unsupported conclusions, incorrect theorem use, missing key geometric conditions, or invalid algebraic manipulation.
    \item For code evaluation, focus on mathematical validity rather than visual aesthetics.
    \item A rendered diagram should be considered correct if it faithfully represents the required geometric or functional relationship, even if its style differs from the reference.
\end{itemize}

\begin{table}[ht!]
\centering
\small
\begin{tabular}{llc}
\toprule
Evaluator~1 & Evaluator~2 & Spearman \\ \hline
Human & Gemini3-Pro & 0.90 \\
Human & DeepSeek-V4-Flash & 0.82 \\
Human & GPT-5.1 & 0.61 \\
\bottomrule
\end{tabular}
\caption{Spearman correlation between human evaluators and LLM evaluators on the textual reasoning correctness.} \label{tab:human_llm}
\end{table}
\paragraph{Pass/fail conversion.}
For consistency with the automatic filtering pipeline, we additionally convert human scores into binary labels: scores $\geq 3$ are treated as passing, while scores $<3$ are treated as failing. We then report agreement between human pass/fail labels and LLM-based pass/fail decisions.

Table~\ref{tab:human_llm} reports the Spearman correlation between LLM evaluators and human evaluators. We observe a particularly high correlation when using Gemini3-Pro, indicating strong agreement with human judgments. For diagram evaluation, the majority of generated diagrams satisfy the required mathematical and structural constraints.

\paragraph{Purpose.}
This human evaluation is used only to validate the reliability of the automatic scoring pipeline, not to replace the full-scale automatic evaluation. In particular, it verifies whether LLM-based judgments align with expert human assessments on reasoning correctness, code correctness, and text-code consistency.

\subsection{Additional Results} \label{app:res}
\begin{table*}[ht!]
\centering
\begin{tabular}{llllllllll}
\toprule
Model & Algebra & Analytic & Cal. & Trig. & Plane & Solid & Stats. & Trans. & Avg. \\ \hline
GPT5.1 & 0.60 & 0.60 & 0.66 & 0.68 & 0.42 & 0.42 & 0.68 & 0.38 & 0.55 \\
GPT5.2 & 0.60 & 0.62 & 0.80 & 0.74 & 0.50 & 0.48 & \textcolor{black}{\textbf{0.84}} & 0.40 & 0.62 \\
Gemini3 Pro & \textcolor{black}{\textbf{0.70}} & 0.70 & 0.82 & \textcolor{black}{\textbf{0.84}} & \textcolor{black}{\textbf{0.68}} & 0.60 & \textcolor{black}{\textbf{0.84}} & 0.66 & \textcolor{black}{\textbf{0.73}} \\
Gemini3 Flash & \textcolor{black}{\textbf{0.70}} & \textcolor{black}{\textbf{0.72}} & \textcolor{black}{\textbf{0.84}} & \textcolor{black}{\textbf{0.84}} & 0.64 & \textcolor{black}{\textbf{0.62}} & 0.76 & \textcolor{black}{\textbf{0.72}} & \textcolor{black}{\textbf{0.73}} \\ \bottomrule
\end{tabular}
\caption{Chosen of pipeline generator, 100 samples of each category randomly sampled from MathCanvas corpus.} \label{tab:baseline}
\end{table*}
Table \ref{tab:corr1} and \ref{tab:corr2} display the Pearson and Spearman correlation from GPT-5.1 and DeepSeek-V4-Flash, respectively. We can observe that incorporating pick-point and rule-based instruction during evaluation can improve the correlation between Ans Acc and Text.
\begin{table*}[ht!]
\centering
\begin{tabular}{lcccc}
\toprule
MLLM & \multicolumn{2}{c}{Ans Acc-Text} & \multicolumn{2}{c}{Ans Acc-Text Code} \\ \hline
\multicolumn{5}{c}{\textit{without rule instruction}} \\
Qwen3-VL-Ins-8B-SFT & 0.43 & 0.43  & 0.36 & 0.37 \\
Qwen3.5-9B-SFT & 0.46 & 0.47  & 0.37 & 0.38 \\
Qwen3.5-9B-LoRA & 0.47 & 0.47  & 0.30 & 0.29 \\
InternVL3.5-8B-SFT & 0.50 & 0.49  & 0.39 & 0.40 \\ \hline
\multicolumn{5}{c}{\textit{with rule instruction}} \\
Qwen3-VL-Ins-8B-SFT & \textbf{0.61} & \textbf{0.59}  & 0.36 & 0.36 \\
Qwen3.5-9B-SFT & \textbf{0.62} & \textbf{0.62} & 0.34 & 0.35 \\
Qwen3.5-9B-LoRA & \textbf{0.64} & \textbf{0.63} & 0.30 & 0.30 \\
InternVL3.5-8B-SFT & \textbf{0.64} & \textbf{0.61} & 0.37 & 0.38 \\

\bottomrule
\end{tabular}
\caption{GPT-5.1: Pearson (left) and Spearman (right) correlation between different metrics.} \label{tab:corr1}
\end{table*}
\begin{table*}[ht!]
\centering
\begin{tabular}{lcccc}
\toprule
MLLM & \multicolumn{2}{c}{Ans Acc-Text} & \multicolumn{2}{c}{Ans Acc-Text Code} \\ \hline
\multicolumn{5}{c}{\textit{without rule instruction}} \\
Qwen3-VL-Ins-8B-SFT & 0.48 & 0.47  & 0.22 & 0.24 \\
Qwen3.5-9B-SFT & 0.49 & 0.48 & 0.22 & 0.23 \\
Qwen3.5-9B-LoRA & 0.49 & 0.48  & 0.21 & 0.22 \\
InternVL3.5-8B-SFT & 0.47 & 0.48 & 0.24 & 0.22 \\ \hline
\multicolumn{5}{c}{\textit{with rule instruction}} \\
Qwen3-VL-Ins-8B-SFT & \textbf{0.68} & \textbf{0.68} & 0.22 & 0.23 \\
Qwen3.5-9B-SFT & \textbf{0.68} & \textbf{0.67} & 0.23 & 0.23 \\
Qwen3.5-9B-LoRA & \textbf{0.69} & \textbf{0.68} & 0.24 & 0.23 \\
InternVL3.5-8B-SFT & \textbf{0.69} & \textbf{0.68} & 0.22 & 0.23 \\
\bottomrule
\end{tabular}
\caption{DeepSeek-V4-Flash: Pearson (left) and Spearman (right) correlation between different metrics.} \label{tab:corr2}
\end{table*}

Table \ref{tab:llm_corr} displays the Spearman correlation between different LLM evaluators. We observe that the correlation between Gemini3-Pro and DeepSeek-V4-Flash is consistently higher than their correlations with GPT-5.1, indicating that Gemini3-Pro and DeepSeek-V4-Flash exhibit more aligned evaluation behaviours for intermediate reasoning assessment.

\begin{figure*}[ht!]
  \includegraphics[width=\linewidth]{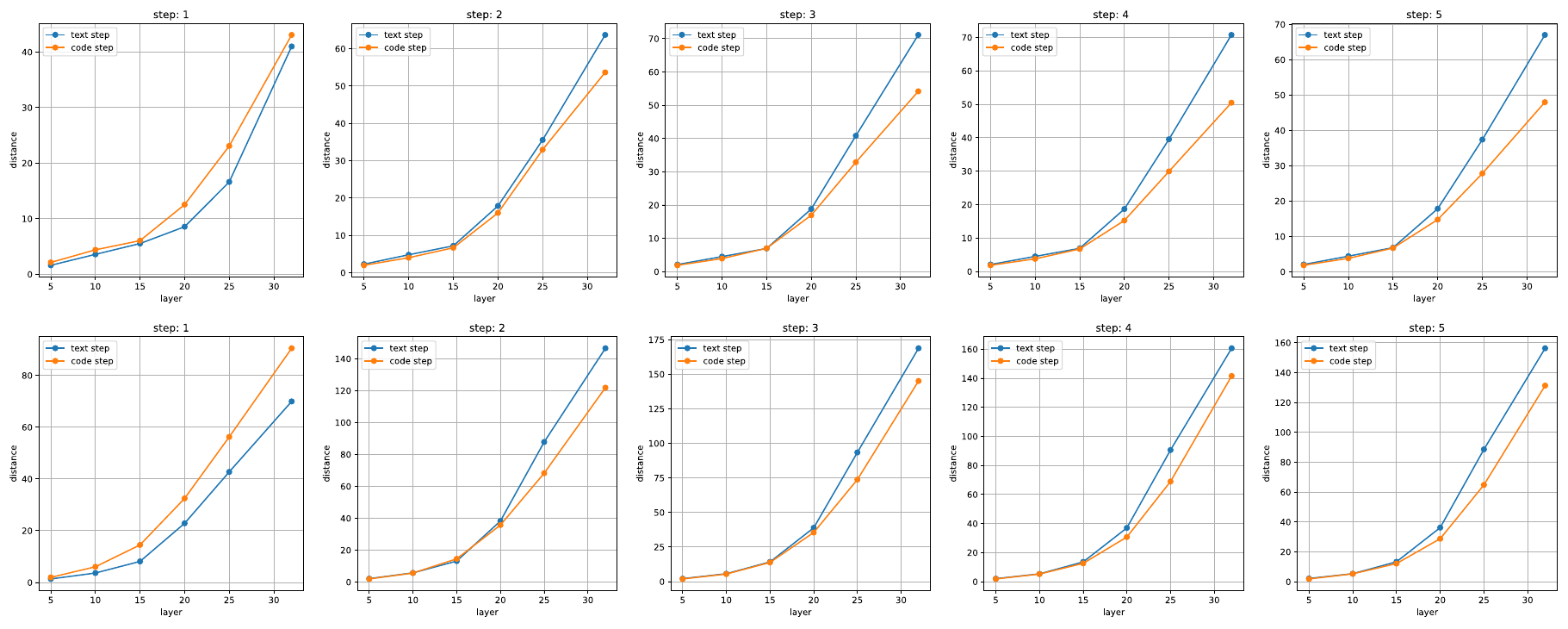}
  \caption{Average euclidean distance of different steps at different layers (Top: Qwen3-ins-8B, bottom: Qwen3-ins-8B-SFT).}
  \label{fig:dist_qwen3}
\end{figure*}
\begin{figure*}[ht!]
  \includegraphics[width=\linewidth]{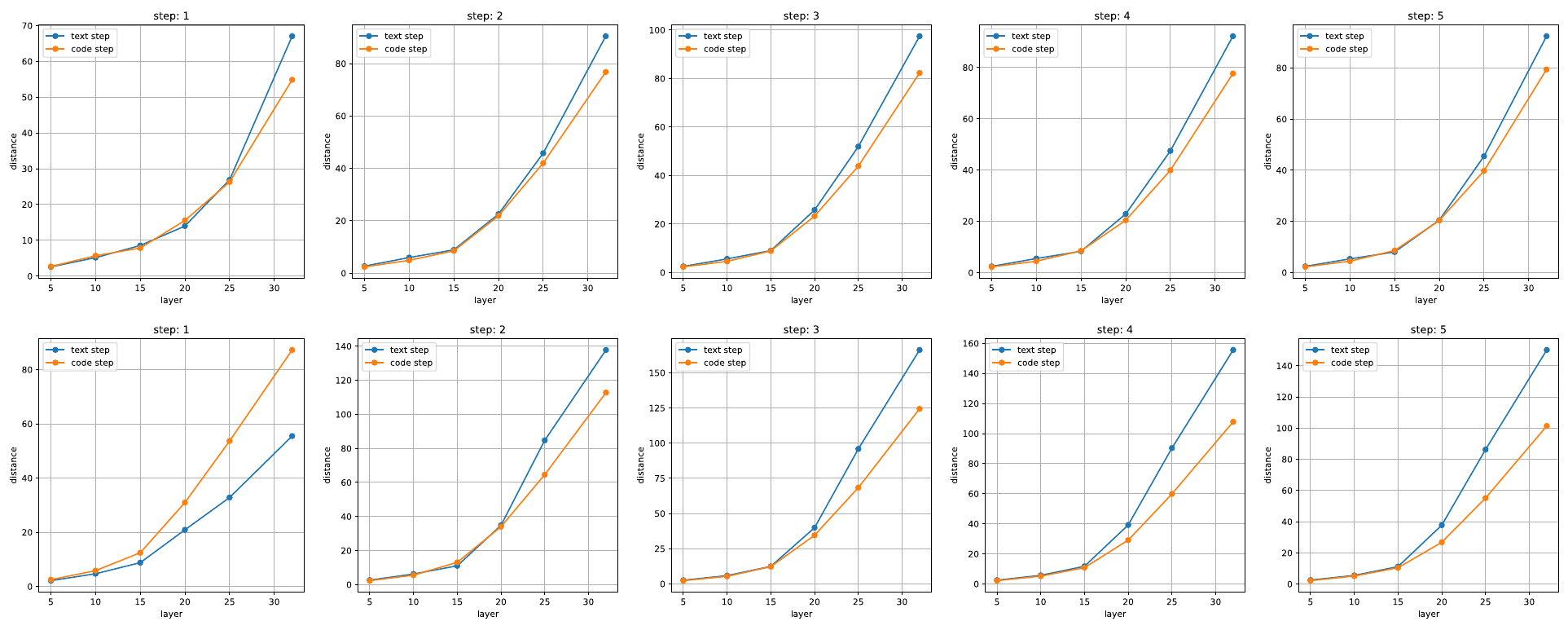}
  \caption{Average euclidean distance of different steps at different layers (Top: InternVL3.5-8B, bottom: InternVL3.5-8B-SFT).}
  \label{fig:dist_internvl}
\end{figure*}

\end{document}